\documentclass[10pt,twocolumn,letterpaper]{article}

\usepackage{amsmath}
\usepackage{amssymb}
\usepackage[ruled,vlined, noend]{algorithm2e}
\usepackage{xcolor}
\usepackage{cvpr}
\pagecolor{white}
\usepackage{enumitem}

\usepackage{graphicx}
\graphicspath{{figures/}}

\usepackage[pagebackref, breaklinks, colorlinks]{hyperref}

\usepackage[capitalize]{cleveref}
\crefname{section}{Sec.}{Secs.}
\Crefname{section}{Section}{Sections}
\Crefname{table}{Table}{Tables}
\crefname{table}{Tab.}{Tabs.}

\begin{document}

\title{SpiderNet: Hybrid Differentiable-Evolutionary Architecture Search via Train-Free Metrics}

\author{Rob Geada\\
Red Hat\\
Newcastle UK\\
{\tt\small rgeada@redhat.com}
\and
Andrew Stephen McGough\\
Newcastle University\\
Newcastle, UK\\
{\tt\small stephen.mcgough@newcastle.ac.uk}
}
\maketitle


\begin{abstract}
	Neural Architecture Search (NAS) algorithms are intended to remove the burden of manual neural network design, and have shown to be capable of designing
    excellent models for a variety of well-known problems. However, these algorithms require a variety of
    design parameters in the form of user configuration or hard-coded decisions which limit the variety of networks
    that can be discovered. This means that NAS algorithms do not eliminate model design tuning, they instead merely shift the burden
    of where that tuning needs to be applied. In this paper\footnote{Published as a workshop paper
in CVPR-NAS 2022.}, we present SpiderNet, a hybrid differentiable-evolutionary and hardware-aware
    algorithm that rapidly and efficiently produces state-of-the-art networks. More importantly, SpiderNet is a proof-of-concept
    of a minimally-configured NAS algorithm; the majority of design choices
    seen in other algorithms are incorporated into SpiderNet's dynamically-evolving search space, minimizing the number
    of user choices to just two: reduction cell count and initial channel count. SpiderNet produces models highly-competitive
    with the state-of-the-art, and outperforms random search in accuracy, runtime, memory size, and parameter count.
\end{abstract}

\section{Introduction}
Neural Architecture Search (NAS) looks to automate network design, to find quick and efficient ways of producing
state-of-the-art networks for novel problems. Differentiable NAS is a popular subcategory of NAS algorithms, which
makes architectural selection a continuous process and thus can be handled by gradient descent during model training,
and has shown to be a tremendously efficient means of performing NAS as compared to very
expensive evolutionary~\cite{real2018,real2017} or reinforcement-learning~\cite{zoph2017,pham2018} based approaches.

While differentiable NAS algorithms can reliably produce good models, their search flexibility is implicitly limited by
their configuration, that is, the parameters that define the supernet within which candidate models are isolated
from. In algorithms like DARTS~\cite{liu2018}, PC-DARTS~\cite{xu2020}, or ProxylessNAS\cite{cai2018}, the search supernet
and thus the NAS search space is fixed throughout the search space,
therefore constricting the available models based on the particular construction of said supernet.
The design of the supernet is often either hardcoded (such as the node count and degree in DARTS) or left to user configuration
(such as the cell stage counts in ProxylessNAS). In essence, this forces the search algorithm into a specific
\textit{network design space} as described by Radosavovic~\etal\cite{radosavovic2020}, but Radosavovic~\etal\;show that particular
choice of design space can have highly significant effects on the performance of models within the design space population.

Given that the intended use-case of NAS is to automatically find optimal architectures for novel problems with minimal
manual tuning, requiring users to define the search space is counter-productive, especially for novel problems
where no network design best-practices exist. Arguably, NAS algorithms
that rely on user configuration of the supernet are just moving the burden of search one level
higher, replacing architecture search with \textit{architecture-search} search: instead of spending time tweaking which
operation goes where, time is spent tweaking things like per-cell node count or reduction cell placement.

With that question in mind, it's worth looking at algorithms that approach NAS from the opposite direction:
rather than specifying the structure of the initial supernet and using differentiable NAS to find some optimal subnet,
what about algorithms that start at some minimal initial state and dynamically expand the search space as necessary?
By letting the NAS algorithm specify the design of its supernet via this expansion,
the flexibility and ease-of-applicability of NAS could be greatly improved.

While evolutionary NAS algorithms like Regularized Evolution~\cite{real2017} exactly fit this required paradigm, they
are plagued by the immense computational costs required to evaluate each candidate in the gene pool. To rectify this,
we present SpiderNet\footnote{Code at \url{https://github.com/RobGeada/SpiderNet-NTKLRC}} a hybrid evolutionary-differentiable NAS algorithm
that uses the train-free Neural Tangent Kernel-Linear Region Count (NTK-LRC) metric~\cite{chen2021} to rapidly evolve its supernet,
while using conventional differentiable techniques to isolate promising subgraphs within. SpiderNet is a minimally-configured
hardware-aware algorithm that designs highly performant networks, and notably demonstrates an
overlooked domain of NAS' advantage over random search: \textit{efficiency}.

In this paper, we will first cover related work to SpiderNet in Section~\ref{sect:relatedwork}, followed by
the design of SpiderNet in Section~\ref{sect:spiderdesign}. Next, Section~\ref{sect:spider_experiments}
describes the various experiments conducted on SpiderNet, the results of which are detailed in Section~\ref{sect:spider_results}.
Finally, sections~\ref{sect:spider_discussion} and~\ref{sect:spiderconclusion} discuss these results and describe their
ramifications to NAS as a whole.

\section{Related Work}~\label{sect:relatedwork}
Evolutionary NAS was first popularized by Neuroevolution~\cite{real2017}, which suffered from immense computational
costs. Meanwhile, differentiable NAS approaches were introduced by Differentiable Architecture Search (DARTS)~\cite{liu2018},
before being refined by papers like Partially Connected DARTS (PC-DARTS)~\cite{xu2020} and ProxylessNAS~\cite{cai2018}.
Meanwhile, train-free NAS techniques are relatively new, with papers like Training-Free NAS (TENAS)~\cite{chen2021} and
NAS Without Training (NASWOT)~\cite{mellor2020} pioneering algorithms that identify promising architectures without
the need for expensive training operations like backpropagation.

\section{Design}\label{sect:spiderdesign}

\subsection{Models}
The SpiderNet supernet is cell structured, where each cell contains edges and nodes.
Edges consist of parallel operations, initially one of each from the 7 available in the
operation search space: Identity, 3x3 Max Pool, 3x3 Average Pool, 3x3  Separable
Convolutions, 5x5 Separable Convolutions, 3x3 Dilated Convolutions, and 5x5 Dilated Convolutions. These are
chosen to align with the common set of computer-vision operations chosen by algorithms like DARTS~\cite{liu2018} or ProxylessNAS~\cite{cai2018}, but
any set could just as easily be chosen. However, unlike DARTS and
ProxylessNAS, edges need not be singular; they can have any arbitrary mixture of operations within the operational set.

Each node performs tensor summation, to ensure that tensor shapes do not change as the number of inbound edges at a node vary.
Each cell has an input node for each previous cell and the original model input; therefore, the fourth cell has four input nodes
(one for Cell 1, 2, and 3, plus the model input), the third has three, etc. Each cell is
initialized with just its input nodes, one intermediate node, and a single output node. One edge is then placed between
each input and the intermediate node, and then a single edge is placed between the sole intermediate node and the output node.
This configuration is shown in Figure~\ref{fig:spider_multiin_initialization}:
\begin{figure}[ht!]
    \centering
  \includegraphics[width=.75\linewidth, trim={1.5cm 1.5cm 1.5cm 1.5cm}, clip]{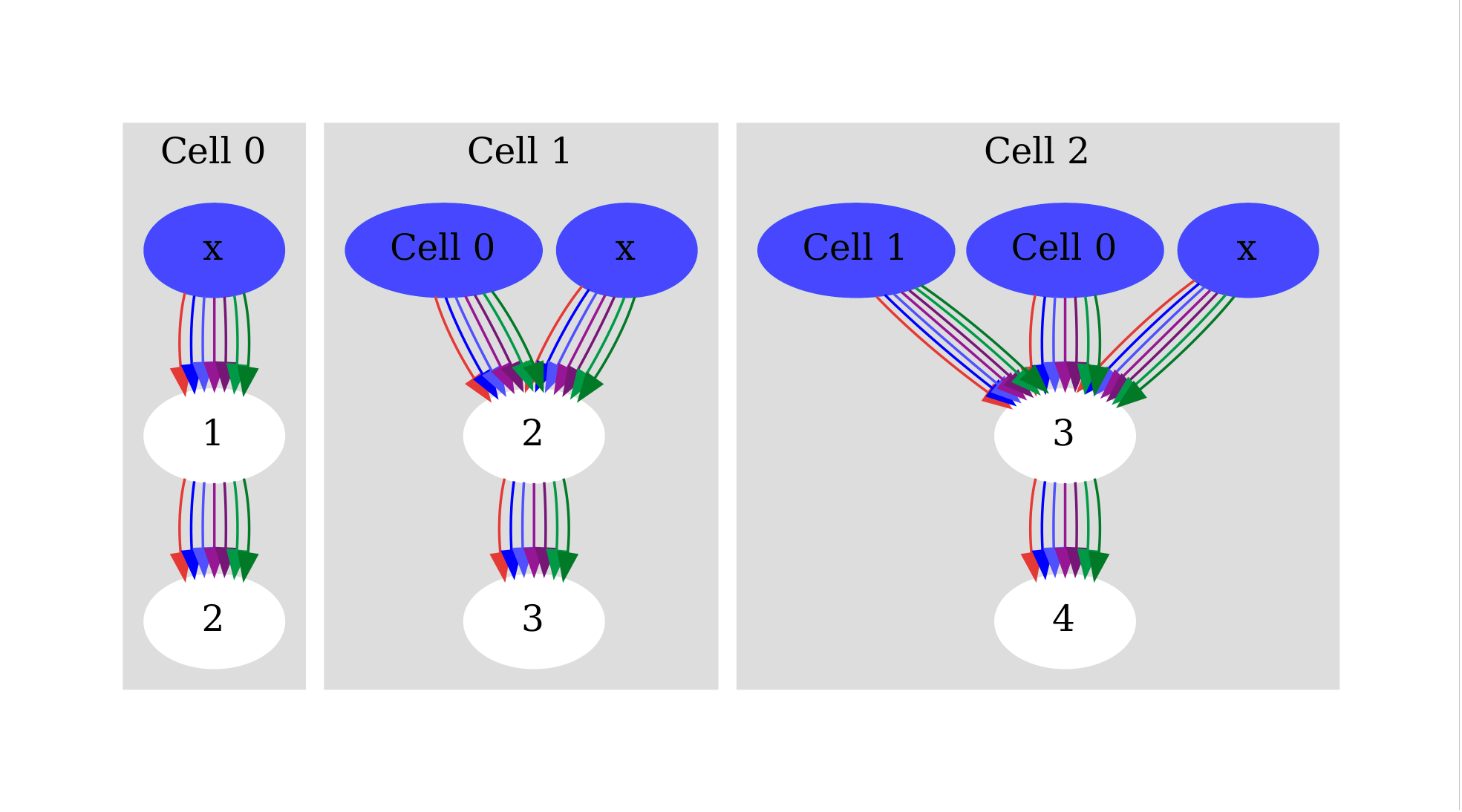}
    \caption[The initial state of a three cell SpiderNet model]{The initial state of a three cell SpiderNet model, with the input nodes marked in blue.}
    \label{fig:spider_multiin_initialization}
\end{figure}

This starting point can be thought of
as a suitably `minimal' starting point as it provides two edges per input: one to the single intermediate and one to the
output. Meanwhile, the single intermediate has an edge to the output that combines and synthesizes information from the inputs.
Evolutions of this architecture can therefore reinforce the pathways from the input or strengthen the `mixing' portion of the cell.

On a macro-structural level, SpiderNet takes inspiration from Xie~\etal\cite{xi2019}, who use a \textit{large-cell} paradigm
for their model design, referring to models that use a small number of high-node cells, as opposed to the high cell-count
low-node cells seen in \textit{small-cell} paradigms like those used by DARTS~\cite{liu2018} or ProxylessNAS~\cite{cai2018}.
The rationale for this is that any cellular pattern comprised of a single reduction cell followed by any number
of sequential `normal' cells (using the standard definition of `normal' and `reduction' cells introduced by Zoph~\etal\cite{zoph_sir2017})
can be replaced as a single reduction cell of sufficiently large node count, while any repeated number of normal cells
can be replaced by a single sufficiently large normal cell. For example, the DARTS model used for CIFAR-10 has 20 small
cells, with reduction cells at positions 7 and 14. This structure can be divided into three groups of cells
\texttt{nnnnnn-rnnnnnn-rnnnnnn}\footnote{Here \texttt{n} and \texttt{r} represent normal and reduction cells respectively.}, which by the above rules reduces to three large cells \texttt{NRR}. The advantage of this
is that it eliminates all cellular design choices beyond reduction count; any possible sequence of small-cells with $r$ reductions
can be represented by $r+1$ large cells: one large normal cell followed by $r$ large reduction cells.

Like in Cai~\etal\cite{cai2018} and Xie~\etal\cite{xi2019}, SpiderNet cells are unique throughout the model; each cell has a unique node count,
edge connectivity, and each edge has a unique operational mixture.

\subsection{Pruning}
To identify promising subnetworks within the supernet, SpiderNet makes use of differentiable pruning, introduced by Kim~\etal\cite{kim2019v2}
and first used for NAS in Geada~\etal\cite{geada2020}. Here, a weighted operation is placed along every edge we wish to make prunable,
that performs the following operation over the inbound tensor $x$:
\begin{align}
    P(x, w) &= \left(Gate(w) + Saw(w)\right)x &\\
    Gate(w) &= \begin{cases}
               0 & w < 0 \\
               1 & w \ge 0
        \end{cases} \\
    Saw(w) &= \frac{Mw - \lfloor Mw \rfloor}{M}
\end{align}
where $M$ is some large constant, in this case $10^9$. This provides a function with the mathematical properties
of a binary gate, but with a constant gradient of 1 with respect to the pruner weight $w$. This allows for $w$ to be
easily learned, meaning operations can
be dynamically pruned during gradient descent to benefit task performance. We can then use this behavior to isolate promising
subgraphs of the SpiderNet supernet.

Our usage of pruners aligns with Geada~\etal\cite{geada2020}, where differentiable pruners are placed along each
parallel edge operation. As the model trains, the pruner weights adapt and switch operations on and off as necessary.
Any pruner which remains switched off for more than 75\% of the batches from the previous 4 epochs is \textit{deadheaded},
that is, permanently deleted from the model. This `locks-in' the pruner decision, freeing up memory space from the unneeded
operation which we can use for subsequent mutations of the network.

\subsection{Mutation}
\begin{figure}[ht!]
\centering
\begin{subfigure}{.45\linewidth}
  \centering
  \includegraphics[height=5em]{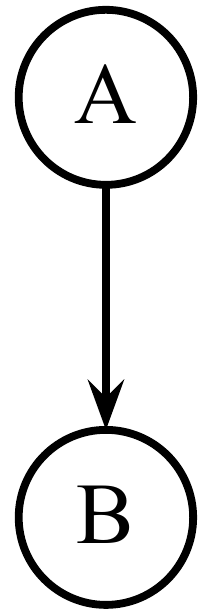}
  \caption{Original Edge}
\end{subfigure}
\begin{subfigure}{.45\linewidth}
  \centering
  \includegraphics[height=5em]{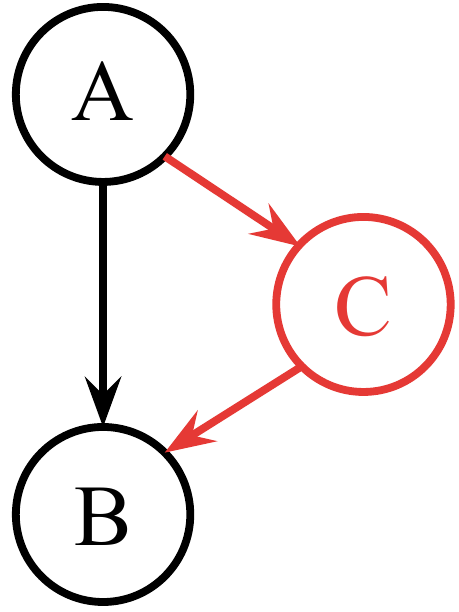}
  \caption{Post mutation}
\end{subfigure}%
\caption[The triangular mutation]{The triangular mutation, with new edges and nodes depicted in red.}
\label{fig:mutation patterns}
\end{figure}

Dynamically expanding the search space is accomplished via \textit{triangular mutations}, which acts on an arbitrary edge $A\rightarrow B$ that
connects two nodes $A$ and $B$ within a model, adding a third node $C$ and two new edges. These edges are wired such that
for the nodes $A$, $B$, and $C$, one node sends two outputs, one node receives one input and sends one output,
and one node receives two inputs, as shown in Figure~\ref{fig:mutation patterns}. Each new edge is initialized with every
operation from the available operation set.
\noindent The triangular mutation
is composable with edge pruning to create an effectively infinite variety of graph patterns, since it crucially introduces a branching
path into a linear node relationship. This creates a seed for further branching by later
mutations along any of these three edges, which allows a small number of mutations to create immense cellular
complexity\footnote{The cells split and grow much like the branching propagation of a spider
plant, hence the name SpiderNet}.
For architectural examples of models generated by such triangular mutations, Figures 1, 2, and 3 in the
Supplementary Material show the final architectures of three models after 45 triangular mutations were
performed.

\subsection{Mutation Practicalities}\label{sect:spider_mutation_reinit}
When mutations are performed during model training, they alter the distributions that downstream operations
are expecting, worsening the model's performance. In reaction to this, the differentiable pruners would instantly
prune away the new operations, thus restoring the connectivity back to the model's pre-mutation state.
To remedy this, the model weights are reinitialized after each mutation, which allows mutation to be a productive
operation.

\subsection{Selecting Edges to Mutate}
With the edge mutation operation defined, the next step is to identify the mutation criteria; how should the model
select an edge to perform this mutation on? To perform this both rapidly and intelligently,
we adapt NTK-LRC to guide mutations.

\subsection{NTK-LRC Train-Free Metric}
A train-free metric refers to methods that evaluate a model's quality via sampling methods that involve
no backpropagation or weight sampling, ideally in such a way that the metric correlates strongly with final
model performance. Following Chen~\etal\cite{chen2021}, we use the neural tangent kernel (NTK) and linear region count
(LRC) to measure a potential architecture's
\textit{trainability} and \textit{expressability} respectively, where trainability refers to how rapidly an architecture trains to good parameter values,
while expressability describes the flexibility of the range of functions the architecture can approximate. Like Chen~\etal,
we use the joint NTK-LRC metric which jointly represents optimal NTK and LRC via the summation of
descending NTK rank to ascending LRC rank as compared to other models within the search population.

\subsection{Searching via Joint NTK-LRC Metric} \label{sect:ntk_metric}
While in theory the NTK-LRC metric allows for every candidate in the search space to be evaluated `for-free' in terms of model
training, the actual computation of the two metrics still comes with a non-trivial computation cost. For larger
search spaces the majority of the space cannot be fully enumerated, and must be traversed in some intelligent manner.
In the case of the DARTS space, Chen~\etal uses a tournament selection~\cite{miller1995} algorithm to find good architectures.
Here, a `champion' model $M$ is compared against a set of challenger models $\mathcal{M}$, each $M'\in\mathcal{M}$ created
by removing one operation at random from $M$. The model $M'\in\mathcal{M}$ with the best joint NTK-LRC is then chosen as
the next champion, and the process repeats until a satisfactory model is found. In essence, this is a train-free method
of iteratively pruning a supernet to find an optimal subnet, and from this approach Chen~\etal find state-of-the-art
results comparable to those found by DARTS.

We adapt Chen~\etal's tournament selection to measure the quality of potential mutations. Given some
model $M$, the set of all models $\mathcal{M}$ that are one mutation away from M is enumerated. The NTK and LRC of
each mutated model $M'$ is compared against the original unmutated model $M$, and the model $M' \in \mathcal{M}$ is selected
that meets the following conditions, if any such model exists:
\begin{enumerate}
    \item Is non-detrimental to both NTK and LRC as compared to $M$
    \item Maximizes the combined NTK-descending and LRC-ascending ranks compared to all other models in $\mathcal{M}$
\end{enumerate}
This process can then be repeated a number of times, in theory improving the quality of the model after each cycle.

One practical caveat to this approach is the aforementioned cost of evaluating NTK and LRC; if each takes around a
minute to evaluate each potential mutation, exhaustively comparing every possible mutation can take around an hour for
larger models. To avoid this, mutation candidates are instead sampled at random, and after
a certain configurable number of `good' mutations are found (i.e., ones that do not increase NTK and do not decrease LRC,
therefore producing a model that is at worst equivalent to the original model) the sampling ends. Of these found
good mutations, the one with the best NTK-LRC is performed.

With this, NTK-LRC can now be used as a mutation metric, and Algorithm~\ref{alg:ntklrc} describes how the NTK-LRC mutation
metric selects candidate edges for mutation.

\begin{algorithm}[ht]
\SetAlgoLined
\SetKwInOut{Input}{Input}
\SetKwInOut{Output}{Output}
\Input{Some model $M$ and sample size $n_{good}$}
Let $s(x)$ refer to the VRAM size of some component $x$\;
Let $s_{max}$ be the maximum permitted VRAM allocation\;
\lnl{ntk:slim} Create $S$, the slim model of $M$\;
Let $\mathcal{E}=\varnothing$\;
\lnl{ntk:loop}\For{\upshape edge $e$ in $M$, selected in random order}{
    \lnl{ntk:copying} Let $S'$ be an exact copy of $S$\;
    \lnl{ntk:mutate} Mutate $e$ in $S'$, thus creating two new edges $e'$ and $e''$\;
    \lnl{ntk:copy_again} Create an identical copy of $S'$ and label the two $S_{on}'$ and $S_{off}'$\;
    \lnl{ntk:disconnect} Within $S_{off}$, disconnect $e'$ and $e''$\;
    \lnl{ntk:compute} Compute NTK and LRC for $S_{off}$ and $S_{on}$\;
    \lnl{ntk:add}\If{\upshape NTK$_{on}$ $\le$ NTK$_{off}$ and LRC$_{on}$ $\ge$ LRC$_{off}$}{
        Add $e$ into the set of good mutation candidates $\mathcal{E}$\;
    }
    \lnl{ntk:break} \If{$|\mathcal{E}| \ge n_{good}$}{
        break\;
    }
}
\If{$|\mathcal{E}| > 0$}{
    \lnl{ntk:findbest} Let $e^*$ be the candidate mutation with best joint NTK-LRC in $\mathcal{E}$\;
    \lnl{ntk:iffits}\If{$s(M) + 2s(e^*) < s_{max}$}{
        \Output{$e^*$}
    }
}
\Output{$\varnothing$}
\caption{NTK-LRC Mutation Selection}
\label{alg:ntklrc}
\end{algorithm}

\begin{figure}[h]
\centering
\begin{subfigure}{.3\linewidth}
  \centering
  \includegraphics[height=5em]{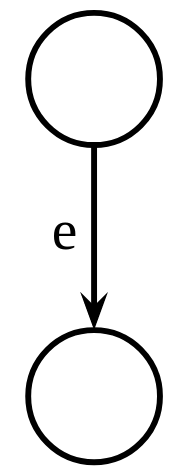}
  \caption{$S$}
\end{subfigure}
\begin{subfigure}{.3\linewidth}
  \centering
  \includegraphics[height=5em]{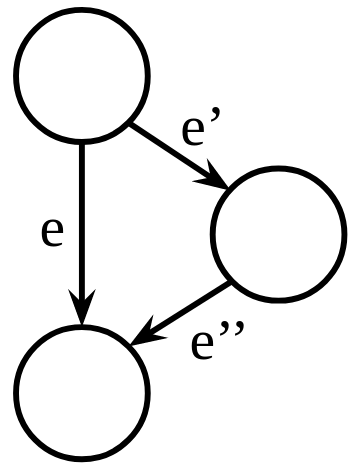}
  \caption{$S_{on}'$, mutated $e$}
\end{subfigure}%
\begin{subfigure}{.3\linewidth}
  \centering
  \includegraphics[height=5em]{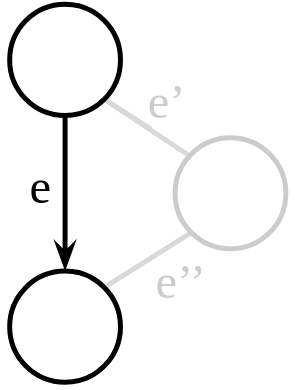}
  \caption{$S_{off}'$, $e'$ and $e''$ disconnected}
\end{subfigure}%
\caption[The state of $S$, $S_{on}'$ and $S_{off}'$]{$S$, $S_{on}'$ and $S_{off}'$, shown in the local area around the
candidate edge $e$. Notice that while $S_{on}'$ and $S_{off}'$ are operationally identical,
(that is, contain identical copies of $e, e'$ and $e''$ and all component operations within), $S_{off}'$ is mathematically
identical to $S$.}
\label{fig:ntk_s_models}
\end{figure}

The NTK-LRC algorithm works by first creating a slim model $S$ of our SpiderNet model $M$, as described in Chen~\etal.
This is done by initializing an exact copy of the SpiderNet model,
and then setting the channel count of each operation to 1. Next, line~\ref{ntk:loop} selects edges at random from the model
as potential mutation candidates. Before trialling the mutation of some selected edge $e$, line~\ref{ntk:copying} creates
an identical copy $S'$ of our slim model $S$; all trial mutations are performed on the copy, thus keeping
the original slim model $S$ as an unmodified template that can be used again. Edge $e$ is then mutated within $S'$ in line~\ref{ntk:mutate},
thus creating two new edges $e'$ and $e''$ as per the rules of triangular mutation. Next, a copy of the mutated $S'$ is made in
line~\ref{ntk:copy_again}. This creates two identical mutated models, the original $S'$ and the copy, which
are then labeled as $S_{on}'$ and $S_{off}'$. Within $S_{off}'$, the new edges $e'$ and $e''$ created by the mutation are
disconnected (simply by multiplying their outputs by 0); this means that $S_{off}'$ is now mathematically identical
to the original unmutated model $S$, but physically has the same parameters as $S_{on}'$. See Figure~\ref{fig:ntk_s_models}
for a simplified visualization of $S$, $S_{on}'$, and $S_{off}'$ at this stage of the algorithm.

Using $S$, $S_{on}'$, and $S_{off}'$ is necessary because NTK-LRC is only useful as a comparative metric between two identically parametered-models.
Since mutation implicitly modifies a model's operations and therefore
parameters, $S$ and $S_{on}'$ cannot be meaningfully compared. Using $S_{on}'$ and $S_{off}'$ remedies that issue, by
creating two identically-parametered models that mathematically represent a mutated and unmutated model respectively. As such,
the NTK-LRC delta induced by the mutation can then be measured, which happens in line~\ref{ntk:compute}.

If this mutation produces a favorable NTK-LRC (that is, did not increase NTK and did not decrease LRC) it is added to the set of
found good mutations $\mathcal{E}$ in line~\ref{ntk:add}. This whole process repeats until a satisfactory number of good mutations
are found or all edges have been sampled, at which point the algorithm exits the edge sampling loop (line~\ref{ntk:break}).
If any good edges are found, then the one with the most favorable joint NTK-LRC is selected in line~\ref{ntk:findbest}
as the final mutation choice. If this mutation is performable on $M$ within the available memory space
(line~\ref{ntk:iffits}), it is returned as the selected mutation. Otherwise, no edge is returned.

\subsection{SpiderNet Algorithm}\label{sect:spider_alg}
With the mutation guiding NTK-LRC metric described, all necessary components for the full SpiderNet algorithm are present.
This is stated in Algorithm~\ref{alg:spider}.

The algorithm first starts by initializing a minimum viable model $M$. Next,
the model is trained and pruned, after which a number of mutations
are performed (if possible), then the whole process repeats until the model has reached sufficient size. Then the full model
trains and prunes freely until it is fully trained.

\begin{algorithm}[ht]
\SetAlgoLined
Initialize `minimum viable model` $M$ as shown in Figure~\ref{fig:spider_multiin_initialization}\;
Let $s(x)$ refer to the VRAM size of some component $x$\;
Let $s_{max}$ be the maximum permitted VRAM allocation\;
Let $E_{train}$ refer to the desired number of training epochs\;
\BlankLine
\For{i in $[1, \#_{mutation\;cycles}$]}{
    \If{\upshape inter-cycle training+pruning is required/desired}{
        Reset model parameters\;
        \For {Epoch in $[1, \#_{epochs\;per\;cycle}]$}{
            \lnl{spider_alg:inter-cycle-pruning} Train + prune\;
            \lnl{spider_alg:deadhead} Deadhead\;
        }
    }
    \For{\# desired mutations}{
        \lnl{spider_alg:mut_metric} Mutate edge returned by Algorithm~\ref{alg:ntklrc} \;
        \If{\upshape no viable mutations}{
            break\;
        }
    }
}
\lnl{spider_alg:intra-train-pruning} Train + prune freely for the $E_{train}$ epochs\;
\caption{The SpiderNet Algorithm}
\label{alg:spider}
\end{algorithm}

\section{Experiment Designs}\label{sect:spider_experiments}
Given the above algorithm and metrics, a variety of experiments can be designed over CIFAR-10 to evaluate their various efficacies,
the details of which are presented in this section.

\subsection{Model and Training Configuration}
Each SpiderNet model has a few configurable hyperparameters, and each model presented in this section
adheres to the following design configuration:

\begin{itemize}   \setlength\itemsep{-.2em}
	\item \textbf{Maximum Permitted VRAM Space}: 20, corresponding to max space of a NVidia 3090.
	\item \textbf{Initial Convolution Channels}: 64
	\item \textbf{Number of Reductions}: 2
	\item \textbf{Operation Set}: Same as DARTS: Identity, 3x3 Max Pool, 3x3 Average Pool, 3x3 Separable Convolution, 5x5 Separable Convolution,
	3x3 Dilated Convolution, 5x5 Dilated Convolution
\end{itemize}

This is the advantage provided by SpiderNet's large-cell structure and dynamically-evolving search space: just two
hyperparameters dictate the search space, initial convolution channels and total reduction count. The other two, VRAM space
and operational choice, are dictated by hardware/deployment constraints and the domain of the given task respectively. Compare this
to something like DARTS, which requires additional choices of total cell count, reduction spacing,
per-cell node choice, node degree, number of input nodes per cell, and number of skip-connections to subsequent cells
to define the search space. SpiderNet handles all of these automatically, dynamically adapting to best fit the task at hand.

In terms of training, all models use the following hyperparameters:
\begin{itemize}   \setlength\itemsep{-.2em}
    \item \textbf{Epochs}: 600
     \item \textbf{Batch Size}: 64
	\item \textbf{Learning Rate}: .01, cosine annealed to 0
	\item \textbf{Dropout}: 0.2
    \item \textbf{Data Augmentation}: Random Cropping, Horizontal Flipping, Cutout, Normalization
\end{itemize}

\subsection{Evolution via Joint NTK-LRC Metric}\label{sect:ntklrc-evolution}
Models are trained and pruned in four epoch cycles. After each cycle, NTK-LRC is used to find up to three beneficial
mutations to perform on the model, provided there is available VRAM space. The model's weights are reset, then the process
repeats. 15 total cycles of this are performed, after which the model trains and prunes as desired for 600 epochs.

\subsection{Random Search and Ablation Studies}\label{sect:spider-random}
Each SpiderNet run is compared against a random search, to evaluate SpiderNet's ability to find good networks within the search space as per Yu~\etal\cite{yu2019}.
In order to do this, it is first necessary to enumerate the different `non-random' components of the algorithm. In the case of SpiderNet, there are
three: inter-cycle pruning, guided mutation by metric, and intra-training pruning, from
lines~\ref{spider_alg:inter-cycle-pruning},~\ref{spider_alg:mut_metric}, and~\ref{spider_alg:intra-train-pruning}
of Algorithm~\ref{alg:spider} respectively.

Replacing guided mutation with a random measure is simple; rather than choose the edges with optimal metric values for
mutation, an edge is selected at random. Each random run then attempts to perform as many random mutations as was attempted
during the compared SpiderNet run, typically 45 (15 cycles of three mutations each). Next, inter-cycle pruning is replaced
by random operation deletion, where the number of operations deleted per cycle is taken from the per-cycle deadhead rates
(Algorithm~\ref{alg:spider}, line~\ref{spider_alg:deadhead}) of the
comparison SpiderNet run. Finally, intra-train pruning is replaced via random operation deletion,
deleting an equal number of operations as were deadheaded during the final model training of the comparison SpiderNet run.
The latter two components can also be simply removed (that is, not performing pruning at all), as a simple ablation study.

Four combinations of these random/non-random/removed components were chosen to provide the widest gamut of comparisons,
the specifics of which are shown in Table~\ref{tab:spider_randoms}.
\begin{table}[ht!]
\begin{center}
\begin{tabular}{c|ccc}
    Label    & Mutation          & \parbox{1.6cm}{\centering Inter-cycle \\ Pruning} & \parbox{1.6cm}{\centering Intra-train \\ pruning} \\ \hline
    Random 1 & Random &  Random & Random \\
    Random 2 & Random &  None   & None \\
    Random 3 & Random &  None   & Pruners \\
    Random 4 & Random &  Random & Pruners \\
\end{tabular}
\end{center}
\caption{Details of the chosen random experiments}
\label{tab:spider_randoms}
\end{table}

These four were chosen for their specific comparative properties. Namely, Random 1 is the pure-random run that replaces
each guided component of Spider with its random equivalent, thus
serving as a purely random reference point against which to compare all experiments against. Random 2 is a pruning ablation
study, which in comparison to Random 1 will inform us of the effects of the random pruning performed. Random 3 evaluates how
well guided-pruning can mesh with random mutation. This will be useful as direct comparison against a guided-run, to determine
how much of its performance stems from the guided-mutation versus guided intra-train pruning. Finally, Random 4 evaluates how
random pruning during the mutation stage (which increases the amount of space available for mutations, therefore
leading to an increased number of mutations as more of the 45 opportunities will be successful) might benefit final
performance.

\section{Results}\label{sect:spider_results}
Table~\ref{tab:spider_all_results} compares the results of SpiderNet against the four random methods. Next,
 Table~\ref{tab:spider_comp_performance} shows a comparison between SpiderNet and other similar NAS methods from the
literature. Since the whole point of SpiderNet is to minimize configurational tuning, reported is the first
run of each configuration.

\begin{table}[ht!]
    \small
\begin{center}
\begin{minipage}{\textwidth}
    \renewcommand{\thempfootnote}{\fnsymbol{mpfootnote}}
\begin{tabular}{r|c|c|c|c|c}
    Growth   & Allotted &  Test  & Search  & Total & Params \\
    Strategy &  VRAM     & Acc.  & Time    & Time \\
    \hline
    SpiderNet         & 20       & \textbf{97.13\%} & \textbf{5.4} & 74.9       & 8.8    \\
    SpiderNet         & \textbf{6.5}      & 96.78\%          & 5.8          & 36.2             & \textbf{4.0} \\
    \hline
    Random 1        & 20       & 96.96\%          & -            & 69.9       & 12.9   \\
    Random 2        & 20       & 97.03\%          & -            & 81.4       & 16.8 \\
    Random 2        & 9.5      & 96.78\%          & -            & \textbf{30.5}       & \textbf{4.0}   \\
    Random 3        & 20       & 96.97\%          & -            & 85.4       & 11.1   \\
    Random 4        & 20       & 96.81\%          & -            & 84.7       & 9.1   \\
\end{tabular}
\end{minipage}
\end{center}
\caption{Results of the six SpiderNet and random runs conducted. Total run time
refers to the start-to-finish wall clock time from initialization to fully-trained model. Search time consists
of the wall clock time of Algorithm~\ref{alg:spider} from start until Line~\ref{spider_alg:intra-train-pruning}.
All search and run times reported in this section are given in GPU hours, max VRAM in gigabytes, and parameter counts in millions.}
\label{tab:spider_all_results}
\end{table}

\begin{table}[ht!]
        \small
\begin{center}
\begin{tabular}{r|c|c|c}
 &&& Search\\
Algorithm & Test Acc. & Params & Hours \\
\hline
NAS-Net\cite{zoph2017}                     & 95.53\% 	                    & 7.1   		& 37,755 \\
ENAS Micro-search\cite{pham2018}	        & 96.131\%            		    & 38.0  		& 7.7     \\
ENAS Macro-search\cite{pham2018}		    & 97.11\%            		    & 4.6   		& 14.4    \\
DARTS First-order\cite{liu2018}     	    & 97.00 $\pm$ 0.14\% 		    & 3.3   		& 36      \\
DARTS Second-order\cite{liu2018}      	    & 97.24 $\pm$ 0.09\% 		    & 3.3   		& 96 \\
NAO (w/ weight sharing)\cite{luo2019}      & 97.07\%        	            & \textbf{2.5}  & 7 \\
ProxylessNAS\cite{cai2018}                & \textbf{97.92}\%                       & 5.7           & N/A           \\
PC-DARTS\cite{xu2020}                    & 97.43\%   	                & 3.6           & \textbf{2.4} \\
 \hline
 \hline
SpiderNet                       & 97.13\% & 8.8     & 5.4 \\
Spider Random 1                 & 97.03\% & 12.9    & -- \\
Spider Random 2                 & 96.96\% & 16.8    & -- \\
Spider Random 3                 & 96.97\% & 11.1    & -- \\
Spider Random 4                 & 96.81\% & 9.1      & --
\end{tabular}
\end{center}
\caption{CIFAR-10 statistics for a variety of similar NAS models covered thus far.}
\label{tab:spider_comp_performance}
\end{table}

\section{Discussion}\label{sect:spider_discussion}
First, the overall ramifications of Tables~\ref{tab:spider_all_results} and~\ref{tab:spider_comp_performance}
will be explored. Then, the table will be broken into specific subsets of runs whose comparisons provide interesting insights.

\subsection{Overall Results}
As compared to the literature, SpiderNet produces highly competitive models. While it is outperformed by algorithms
like ProxylessNAS and PC-DARTS, it does so sans the extensive configurational choices of things like node count
and cell structure required by these algorithms. In situations where the best-practices for these choices are unavailable,
these algorithms would require extensive tuning not required by SpiderNet.

Additionally, the runs of lower VRAM specification demonstrate that the SpiderNet performs well under different constraints:
apart from differing VRAM caps, these runs were configured identically. This demonstrates a versatility for performing
NAS in different hardware domains, specifically one that comes with very little need for configuration tuning. While
the configurations that performed well in the high-VRAM domain may not be optimal for low-VRAM domains (nor are
they guaranteed to be optimal even in their intended range), SpiderNet finds very good models regardless. This is exciting
from an ``applied NAS'' perspective, in that it minimizes the need to tune the algorithm to the desired domain.
This means that SpiderNet has fulfilled the goals that were outlined in the introduction, wherein
it was of interest to explore an algorithm free of the structural configuration tuning that weighs down fixed-structure algorithms like
DARTS or ProxylessNAS.

In terms of the randomly mutated models, the non-pruned Random 2 produced the best performance,
while the pure-random Random 1 produced the fastest model by runtime. This is an interesting result that ran
contrary to expectations, which stemmed from prior differentiable pruning results~\cite{geada2020,kim2019v2}. In these,
differentiable pruning during training was shown to increase accuracy and decrease cost as compared to random pruning or no
pruning at all.

From these expectations, the results of Random 2 and Random 3 are surprising; if the unpruned
supernet of Random 2 performed so well, it might be expected that a similarly-grown supernet that could differentiably prune would again
produce better accuracy more cheaply. However, Random 3 produced worse accuracy four hours slower than Random 2, although it
did use 5.7M fewer parameters. The same could be said about the Random 1 and Random 4 pair; we might expect Random 4
to perform better due to differentiably pruning during training. However, Random 4 was significantly less accurate than
Random 1 as well as around 15 hours slower in total runtime.

One possible explanation to these results is that all previous pruner experiments were performed over identical
supernets, while these four random runs each have a different randomly generated structure. Given how drastically
the structure of SpiderNet models can vary (see Figures 1, 2, and 3 in the
Supplementary Materials for examples of this), the
performance differences seen may be the result of the architecture's implicit capabilities, more so than the effects
of pruning. To fully decouple these effects, the random methods would all need to be performed over the same
randomly-mutated model, that is, pick a random sequence of mutations that are then deterministically performed
every time. However, this is a tangential line of experimentation to the main focus of the work, and thus was not
prioritized.

\subsection{Best NAS versus Best Random}\label{sect:spider_bestvbest}
The first analyzed pairing is the best SpiderNet  run against the best random run, shown in Table~\ref{tab:spider_rand2}:

\begin{table}[ht!]
    \small
\begin{center}
\begin{tabular}{r|c|c|c|c|c}
    Growth   & Allotted &  Test   & Search& Total & Params \\
    Strategy &  VRAM     & Acc.  &  Time    & Time \\
    \hline
    SpiderNet         & 20       & \textbf{97.13\%} & 5.4          & \textbf{74.9}       & \textbf{8.8}    \\
    Random 2        & 20       & 97.03\%          & -            & 81.4       & 16.8 \\
\end{tabular}
\end{center}
\caption{The best random run (Random 2) versus the best SpiderNet run.}
\label{tab:spider_rand2}
\end{table}

The most immediate conclusion that may be drawn from this is that there are only marginal accuracy differences between
the SpiderNet and random runs, with the best SpiderNet run scoring only 0.10 percentage points better than the best random run.
However, there is a large parameter difference between the two; the SpiderNet run used less than half the parameters of the
random run. Additionally, the SpiderNet run from start to finish took around six hours less than the random run. These latter
two points are particularly interesting; the whole random-search condemnation of NAS stems from random search providing
a model of equivalent accuracy and parameter count ``for free''. In any search space wherein every candidate model
has an identical count of single-path edges (DARTS or ProxylessNAS, for example) this
might be a valid assumption, as the single-operation-per-edge restriction means that every
possible model contains an identical number of operations (equal to the total number of edges $e$ in the model),
meaning training times will only vary based on the computational complexity of the selected operations. Meanwhile, if
operations (and thus per-edge parameter counts) are selected from a uniform distribution (such as in~\cite{li2019}),
it is expected that the total number of parameters in any randomly generated model will also be roughly similar.\footnote{
To test this, 100,000 models were generated from a single-path, 57 edge search space via uniform operation selection.
    The distribution of parameter counts was highly normal (Shapiro-Wilk statistic 0.9996, p=$4.1\times 10^{-12}$),
    with a mean of 1.67M and standard deviation of 0.28M.} As a result, neither the Yu~\etal\cite{yu2019} nor Li \&  Talwalkar~\cite{li2019} papers
that started the random search discussion consider model training time a factor, and only the
latter mentions parameter count, albeit only to point out the equivalent parameter counts of their random and NAS models.

Meanwhile, the SpiderNet search space is entirely unbounded; models with
only three operations are valid members of the search space as are those with 300 or 3,000. Therefore, the time costs of training and parameter
counts of any such model in the space varies wildly, and is a massively important factor in determining the quality of any given model. In this
regard, the SpiderNet model's speed is notable; not only does it provide an equivalently accurate (in fact,
better) model than the best random-search model, this model is so parameter and time-efficient that its 5.4 hour
search time handicap is more than doubly made up over the course of training. Additionally, the SpiderNet run only used around
15 GB of its maximum allotted 20 GB, while Random 2 used the full 20 GB, another efficiency not considered in random-search studies.

\subsection{Best NAS versus Pure Random}
Next is the comparison of the best SpiderNet run to the purely random run, shown in Table~\ref{tab:spider_rand1}:
\begin{table}[ht!]
        \small
\begin{center}
\begin{tabular}{r|c|c|c|c|c}
    Growth   & Allotted &  Test   & Search  & Total & Params \\
    Strategy &  VRAM     & Acc.  &  Time    & Time \\
    \hline
    SpiderNet         & 20       & \textbf{97.13\%} & 5.4          & 74.9                 & \textbf{8.8}    \\
    Random 1        & 20       & 96.96\%          & -            & \textbf{69.9}       & 12.9   \\
\end{tabular}
\end{center}
\caption[The fastest random run versus the best SpiderNet run]{The fastest random run (Random 1) versus the best SpiderNet run.}
\label{tab:spider_rand1}
\end{table}

The pure random run compares more favorably to the best SpiderNet run, with a faster training time and closer
parameter count, at the expense of marginally worse accuracy than Random 2. However, it still has around 4 million
more parameters than the SpiderNet run. These results show that random search is relatively competitive
against SpiderNet, albeit with consistently worse parameter efficiency and accuracy.

\subsection{Constricted NAS versus Constricted Random}
\begin{table}[ht!]
        \small
\begin{center}
\begin{tabular}{r|c|c|c|c|c}
    Growth   & Allotted &  Test   & Search  & Total & Params \\
    Strategy &  VRAM     & Acc.  &  Time    & Time \\
    \hline
    SpiderNet         & \textbf{6.5}      & 96.78\%          & 5.8          & 36.2             & \textbf{4.0} \\
    Random 2        & 9.5      & 96.78\%          & -            & \textbf{30.5}    & \textbf{4.0}   \\
\end{tabular}
\end{center}
\caption[The constricted random run versus the constricted SpiderNet runs]{The constricted random run versus the constricted SpiderNet run.}
\label{tab:spider_constrict}
\end{table}
The runs in Table~\ref{tab:spider_constrict} comprise a set of runs where overall, top-end accuracy was less prioritized in favor of
parameter-efficiency (the SpiderNet and Random 2 runs). These two perform nearly identically,
with SpiderNet taking 5.7 hours longer due to its search time
costs but requiring 3 GB less VRAM.

\section{Conclusion}\label{sect:spiderconclusion}
SpiderNet stands as a proof-of-concept of a minimally-configured NAS algorithm, one that aims to minimize
the number of manual choices needed to produce excellent networks. By operating within an unbound, infinite search space
that is specified by just two parameters (reduction count and initial convolution scale), SpiderNet has the capability
to incorporate previously manual design parameters like per-cell node count automatically into its search space, drastically
reducing the available tuning parameters required to produce good models.

Another point of interest amongst our results is the triangular mutation's ability to produce
very high quality networks from minimal starting conditions. In this sense, this work is comparable to that
of Real~\etal\cite{real2018}'s Regularized Evolution, which likewise grew models from minimal conditions
to state-of-the-art capabilities. Indeed, the whole concept of ``minimal initial conditions'' is directly inspired by
the ``trivial initial conditions'' described in Real~\etal\cite{real2018}. However, what is interesting about the triangular mutation
is that it can do this so quickly and consistently,
without much need to carefully consider \textit{where} it should occur. The random results show this
clearly, as although they are all slightly worse in various metrics than the SpiderNet runs, they are still comparable
in performance to various other NAS approaches.

\subsection{The False Equivalence of Random Search}
This strong accuracy performance of the randomly grown models does not invalidate the results of SpiderNet, in fact, the opposite. Crucially,
it revealed that comparison to random search in these idiosyncratic search spaces can create a false-equivalency. If
the entire end-to-end process of finding and training a model takes as long or longer for a random algorithm than
NAS, then any accuracy similarities between the two become significantly less relevant. Section~\ref{sect:spider_bestvbest}
make this very clear; spending just five hours searching is rewarded by increased accuracy, lowered VRAM, and halved
parameter counts, and then the search cost is entirely refunded and more by the increased training speed. This completely
erodes any argument that random search's comparable accuracy detracts from NAS; such an argument entirely
ignores the three other dimensions of parameter, VRAM, and time efficiency by which SpiderNet and NAS show clear benefits.

{\small
\bibliographystyle{ieee_fullname}
\bibliography{bib}

\begin{thebibliography}{10}\itemsep=-1pt

\bibitem{cai2018}
Han Cai, Ligeng Zhu, and Song Han.
\newblock Proxylessnas: Direct neural architecture search on target task and
  hardware.
\newblock {\em CoRR}, abs/1812.00332, 2018.

\bibitem{chen2021}
Wuyang Chen, Xinyu Gong, and Zhangyang Wang.
\newblock Neural architecture search on {ImageNet} in four {GPU} hours: {A}
  theoretically inspired perspective.
\newblock {\em CoRR}, abs/2102.11535, 2021.

\bibitem{geada2020}
Rob Geada, Dennis Prangle, and Andrew~Stephen McGough.
\newblock Bonsai-net: One-shot neural architecture search via differentiable
  pruners.
\newblock {\em CoRR}, abs/2006.09264, 2020.

\bibitem{kim2019v2}
Jaedeok Kim, Chiyoun Park, Hyun-Joo Jung, and Yoonsuck Choe.
\newblock Plug-in, trainable gate for streamlining arbitrary neural networks.
\newblock {\em CoRR}, abs/1904.10921, 2019.

\bibitem{li2019}
Liam Li and Ameet Talwalkar.
\newblock Random search and reproducibility for neural architecture search.
\newblock {\em CoRR}, abs/1902.07638, 2019.

\bibitem{liu2018}
Hanxiao Liu, Karen Simonyan, and Yiming Yang.
\newblock {DARTS:} differentiable architecture search.
\newblock {\em CoRR}, abs/1806.09055, 2018.

\bibitem{luo2019}
Renqian Luo, Fei Tian, Tao Qin, Enhong Chen, and Tie-Yan Liu.
\newblock Neural architecture optimization.
\newblock {\em CoRR}, abs/1808.07233, 2018.

\bibitem{mellor2020}
Joseph~Charles Mellor, Jack Turner, Amos~J. Storkey, and Elliot~J. Crowley.
\newblock Neural architecture search without training.
\newblock {\em CoRR}, abs/2006.04647, 2020.

\bibitem{miller1995}
Brad~L. Miller, Brad~L. Miller, David~E. Goldberg, and David~E. Goldberg.
\newblock Genetic algorithms, tournament selection, and the effects of noise.
\newblock {\em Complex Systems}, 9:193--212, 1995.

\bibitem{pham2018}
Hieu Pham, Melody~Y. Guan, Barret Zoph, Quoc~V. Le, and Jeff Dean.
\newblock Efficient neural architecture search via parameter sharing.
\newblock {\em CoRR}, abs/1802.03268, 2018.

\bibitem{radosavovic2020}
Ilija Radosavovic, Raj~Prateek Kosaraju, Ross~B. Girshick, Kaiming He, and
  Piotr Doll{\'{a}}r.
\newblock Designing network design spaces.
\newblock {\em CoRR}, abs/2003.13678, 2020.

\bibitem{real2018}
Esteban Real, Alok Aggarwal, Yanping Huang, and Quoc~V. Le.
\newblock Regularized evolution for image classifier architecture search.
\newblock {\em CoRR}, abs/1802.01548, 2018.

\bibitem{real2017}
Esteban Real, Sherry Moore, Andrew Selle, Saurabh Saxena, Yutaka~Leon Suematsu,
  Quoc~V. Le, and Alex Kurakin.
\newblock Large-scale evolution of image classifiers.
\newblock {\em CoRR}, abs/1312.4400, 2017.

\bibitem{xi2019}
Saining Xie, Alexander Kirillov, Ross Girshick, and Kaiming He.
\newblock Exploring randomly wired neural networks for image recognition.
\newblock {\em CoRR}, abs/1904.01569, 2020.

\bibitem{xu2020}
Yuhui Xu, Lingxi Xie, Xiaopeng Zhang, Xin Chen, Guo-Jun Qi, and Qi~Tian amd
  Hongkai~Xiong.
\newblock {PC-DARTS:} partial channel connections for memory-efficient
  architecture search.
\newblock {\em CoRR}, abs/1907.05737, 2020.

\bibitem{yu2019}
Kaicheng Yu, Christian Sciuto, Martin Jaggi, Claudiu Musat, and Mathieu
  Salzman.
\newblock Evaluating the search phase of neural architecture search.
\newblock {\em CoRR}, abs/1902.08142, 2019.

\bibitem{zoph2017}
Barret Zoph and Quoc~V. Le.
\newblock Neural architecture search with reinforcement learning.
\newblock {\em arXiv preprint arXiv:1611.01578}, 2017.

\bibitem{zoph_sir2017}
Barret Zoph, Vijay Vasudevan, Jonathon Shlens, and Quoc~V. Le.
\newblock Learning transferable architectures for scalable image recognition.
\newblock {\em CoRR}, abs/1707.07012, 2017.

\end{thebibliography}
}

\end{document}


\maketitle

\vspace{-3em}
\section{Final Model Architectures}
\begin{figure}[ht]
    \centering
    \textbf{Random Model}\par\medskip
	\includegraphics[width=\linewidth, trim={1.5cm 1cm 1.5cm 1cm}, clip]{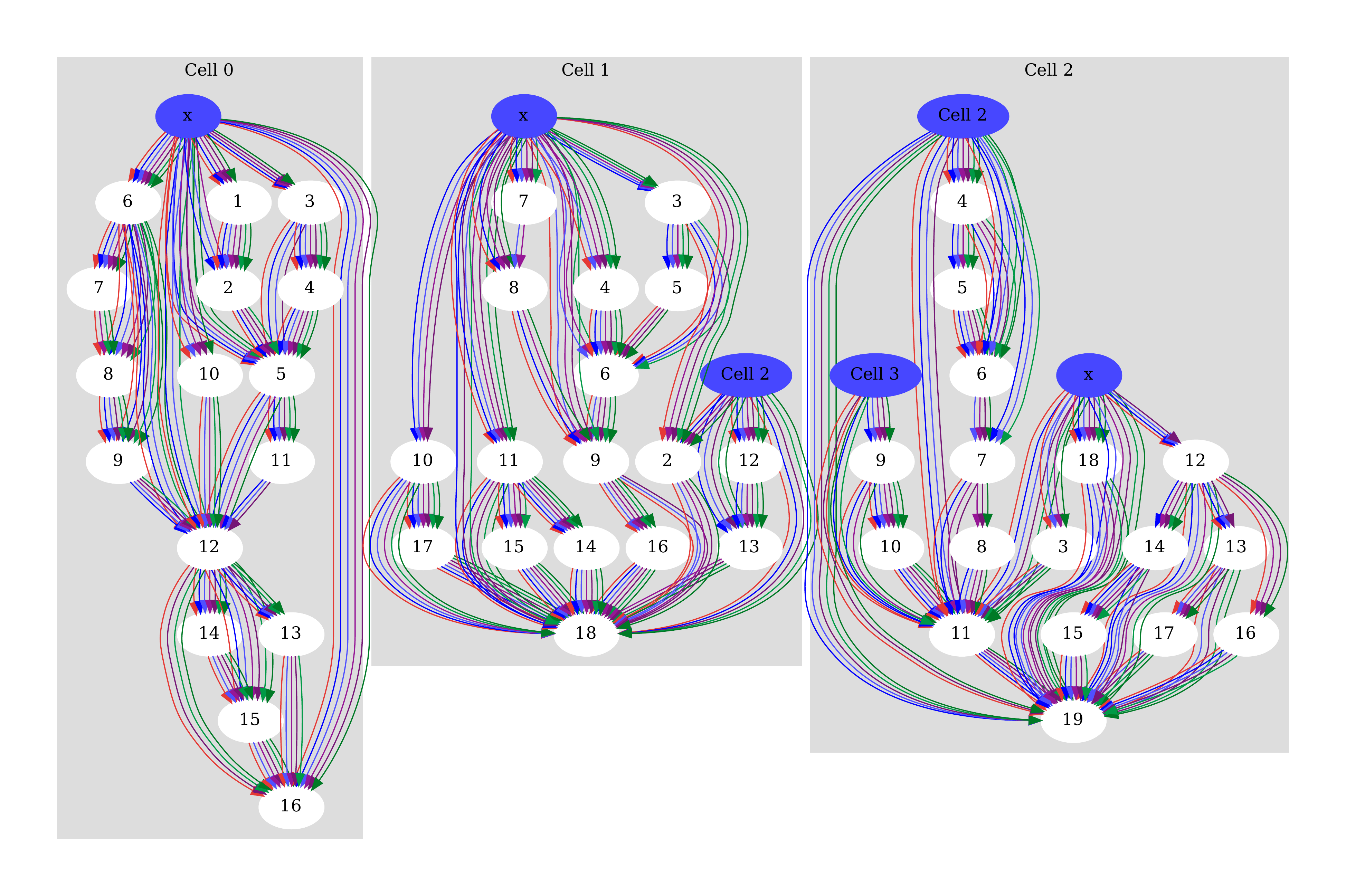} \\
    \caption{A sample SpiderNet model after 45 random mutation attempts and random pruning. Cellular inputs are shown in blue.}
    \label{fig:spider_rand0}
\end{figure}

\begin{figure}[ht]
    \centering
    \textbf{SpiderNet Model}\par\medskip
	\includegraphics[width=\linewidth, trim={1.5cm 1cm 1cm 1cm}, clip]{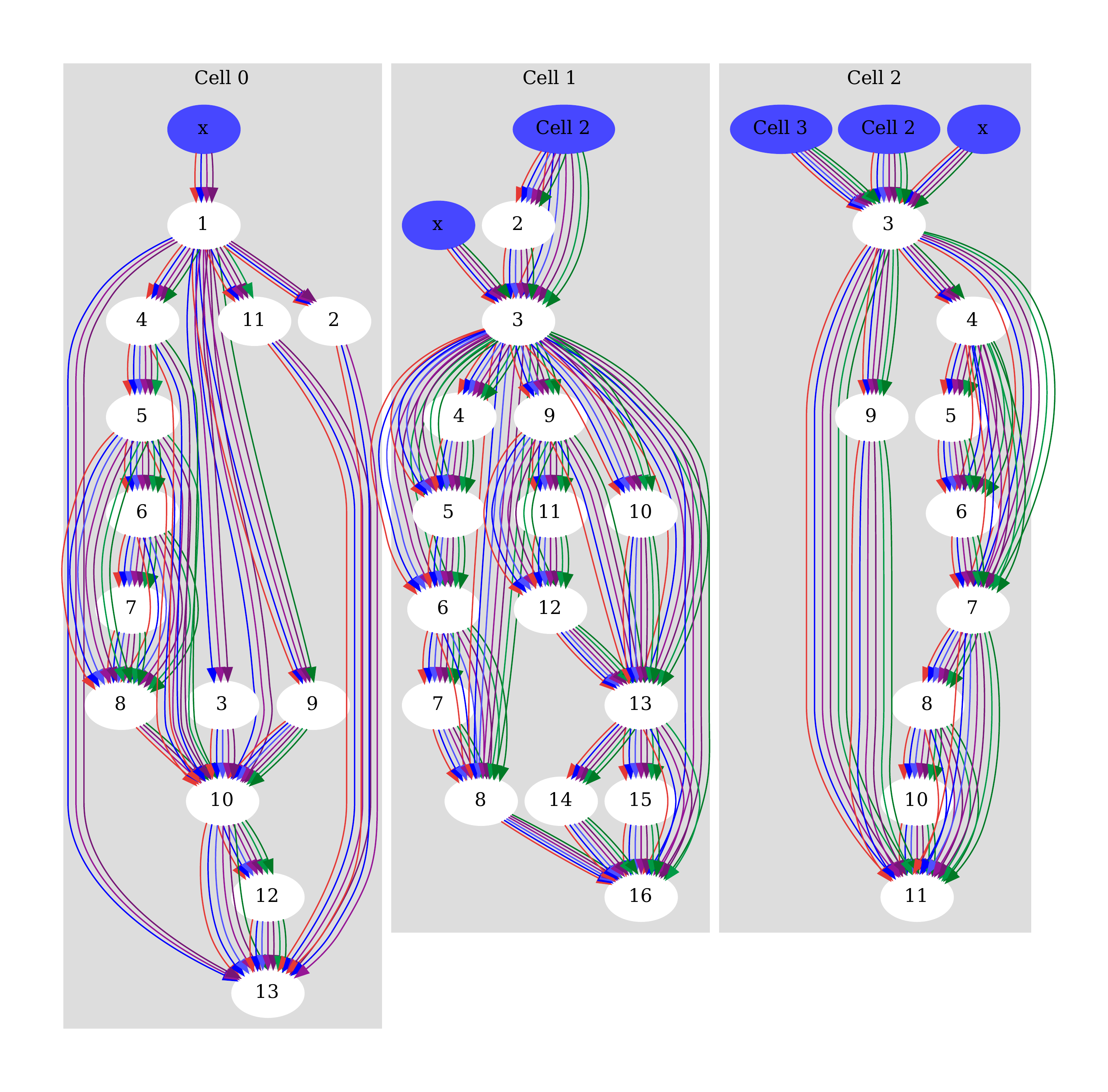} \\
    \caption{A SpiderNet model after 45 NTK-LRC guided mutation attempts and guided inter-cycle pruning. Cellular inputs are shown in blue.}
    \label{fig:spider_ntklrc}
\end{figure}

This section will display examples of the complex architectures produced by the composition of mutations and pruning.
Figure~\ref{fig:spider_rand0} shows a random model created by choosing 45 random mutations and random intercycle pruning. Meanwhile,
Figure~\ref{fig:spider_ntklrc} shows a SpiderNet model created by 45 NTK-LRC guided mutations and inter-cycle pruning via differentiable pruners.
Notice that as compared to the random model, SpiderNet ensures that the input nodes arrive very early into the cell,
never later than the third node. Additionally, SpiderNet produces lower-node count cells than that of the random models despite
having an equivalent number of mutation opportunities; the SpiderNet model has 13, 16, and 11-node cells, while the random
model has 16, 18, and 19-node cells. This is a visual demonstration of the SpiderNet's efficiency, in its ability to
produce superior models to random search in significantly smaller amounts of VRAM space and parameter count.

